\documentclass[10pt,letterpaper]{article}

\usepackage{ccn}
\usepackage{pslatex}
\usepackage{apacite}
\usepackage{subcaption}
\usepackage{graphicx} 
\usepackage{float}
\usepackage[table,xcdraw]{xcolor}
\usepackage[normalem]{ulem}

\usepackage[table]{xcolor} 
\usepackage{booktabs}  
\usepackage{amsmath}   
\usepackage{caption}   
\definecolor{myred}{HTML}{FE0000}

\useunder{\uline}{\ul}{}

\title{Hues and Cues: Human vs. CLIP}

\author{\large \bf Nuria Alabau-Bosque$^{a*}$, \bf Jorge Vila-Tomás$^{a}$,  \bf Paula Daudén-Oliver$^{a}$ , \bf Pablo Hernández-Cámara$^{a}$ \\  \bf Jose Manuel Jaén-Lorites$^{b}$, \bf Valero Laparra$^{a}$, Jesús Malo$^{a}$ \\
$^{a}$ Image Processing Lab, Universidad de Valencia, Paterna, Spain\\
$^{b}$ Center for Biomaterials and Tissue Engineering Universitat Politecnica de Valencia,  Valencia, Spain\\
$^{*}$ Corresponding author: nuria.alabau@uv.es}
 
\begin{document}

\maketitle

\section{Abstract}
{\bf
Playing games is inherently human, and a lot of games are created to challenge different human characteristics. However, these tasks are often left out when evaluating the human-like nature of artificial models. The objective of this work is proposing a new approach to evaluate artificial models via board games. To this effect, we test the color perception and color naming capabilities of CLIP by playing the board game \emph{Hues \& Cues} and assess its alignment with humans. Our experiments show that CLIP is generally well aligned with human observers, but our approach brings to light certain cultural biases and inconsistencies when dealing with different abstraction levels that are hard to identify with other testing strategies. Our findings indicate that assessing models with different tasks like board games can make certain deficiencies in the models stand out in ways that are difficult to test with the commonly used benchmarks.
}
\begin{quote}
\small
\textbf{Keywords:} 
human alignment; model evaluation; perception.
\end{quote}

\section{Introduction}

Board games are often designed with the perceptual capabilities of humans in mind. They can be aimed at presenting challenges in reasoning, perceptual, and ability to players while making the competition a pleasant experience. This makes them an excellent testing field to assess how human-like are modern artificial models in tasks that can vastly differ from the traditional benchmarks commonly used.

There is a wide variety of board games, but \emph{Hues and Cues} is a game that poses a gamification of color-naming in a similar manner as in \cite{brown_color_2023}: players try to match concepts to colors in the board. This simple mechanic puts to test the color perception of the players and sets up a scenario where artificial models can also be tested. Even if the game does not impose language, it can show biases induced by culture and language because it depends broadly on where and with whom is played, as seen in the color-naming literature \cite{lindsey_lexical_2021, kay2023world}.

In this work we showcase how this particular board game (Hues and Cues) can be used to assess the color perception capabilities of Contrastive Language-Image Pretraining (CLIP) model \cite{clip}, a widely used model that is devoted to measuring distances between words and images, and its alignment with humans.

\section{Methods}


Hues and Cues is a competitive party board game for 3 to 10 players. In turns, a player (leader) draws a card with a color and its coordinates. The leader gives a single word clue to guide the other players to the color selected on the card. The rest of the players, in turns, will place their marker on the board with their interpretation of the clue. There is a second round in which the leader can use a concept of up to two words to make the other players rectify, who will place their second and last marker. At the end of these two rounds, a region of $3\times3$ squares centered on the target square is established, and points are calculated according to the number of player markers in the highlighted region region for the leading player and the proximity to the target color for the other players. Figure \ref{fig:Colores} shows a picture of the board, with the scoring region at the top and the playing region in the middle. 
The color distribution is smooth in the chromatic diagram, but saturation and luminance are not taken uniformly.


\subsection{Measures with humans and models}

\begin{figure}[!h]
	\centering
	\includegraphics[width=1\linewidth]{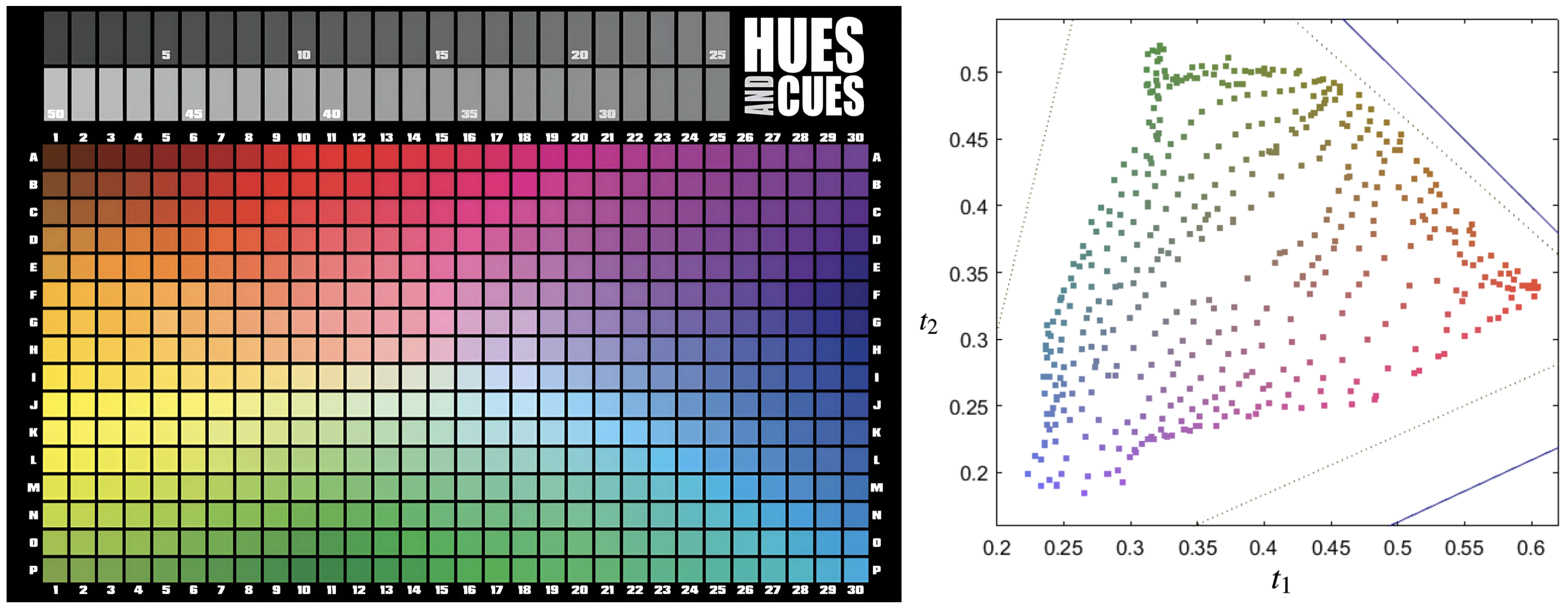}
	\caption{\small{Image of the original board game (left). It has a total of 480 colors organized in 16 rows and 30 columns, and two main directions, red-green and blue-yellow, analogous to opponent color spaces. Data replication in the CIE xy chromatic diagram after measuring the board game with a colorimeter (right).}}
	\label{fig:Colores}
\end{figure}

Since CLIP is a model trained to calculate similarities between text and images, we have simplified the game by reducing it to one color choice per word. In addition, we have pre-selected 34 words to restrict the word-space. To obtain the human measurements we independently ask 6 people which color they would mark for each word. To obtain the model's responses we pass the 480 colors (as flat stimulus) to the CLIP model together with the 34 words. Figure \ref{fig:Colores} shows the original board game (left) and the corresponding CIE xy chromatic coordinates measured with a colorimeter (right). The flat stimuli passed to the model are generated from these measurements~\cite{Colorlab02}. For each word, we collect top-5 more similar colors.

\begin{figure*}[]
	\centering
	\includegraphics[width=0.85\linewidth]{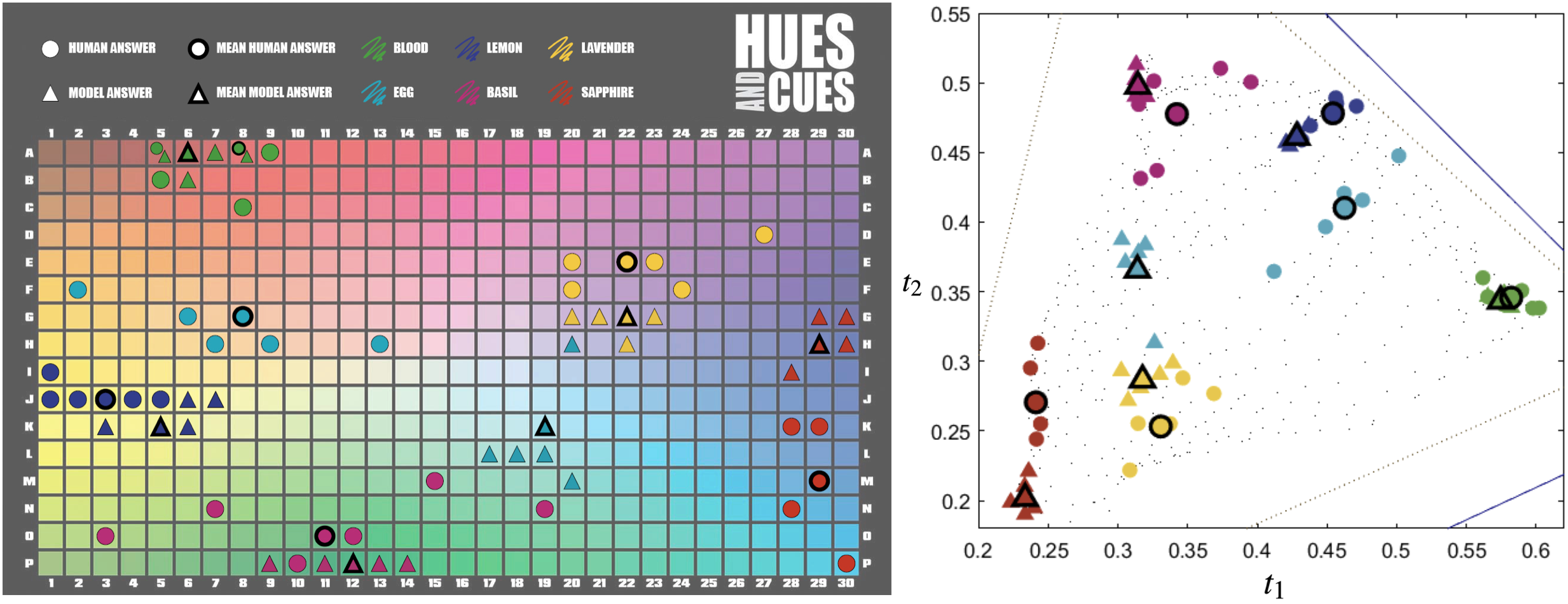}
	\caption{Results of the experiments: on the board game (left), on the chromatic diagram (right). Results for humans, and top 5 model responses for 6 different words are shown. 
    Bold shapes correspond to the mean of the human responses (circles) and the weighted mean of the top 5 model responses (triangles).}
	\label{fig:MediasHC}
    \end{figure*}

\subsection{Evaluation}
\label{sec:evaluation}
In order to evaluate if the responses from the humans and the responses from the model could share the same mean (and thus correspond to the same answer), we employ the Hotelling test \cite{Hotelling}. This test can be understood as a generalization of the t-test to multivariate distributions. We perform this test in the chromatic diagram.
The p-value chosen to discard the null hypothesis (humans and CLIP answering with the same color) is 0.003 ($3\sigma$). Model's responses with bigger p-value are considered to match humans, while lower p-values are considered to do not match. This confidence interval is bigger than the usual $2\sigma$ because it is chosen to match what a human evaluator would consider good answers with respect to the game's criterion, hence being more permissive.

\section{Results}

A representative subset of the measurements obtained from humans (circles) and CLIP (triangles) are shown in Fig. \ref{fig:MediasHC}.
The subset of words represented was chosen to avoid cluttering the image, but results of the Hotelling test for all the words are shown in Table \ref{tab:hotelling}. 
Humans measurements tend to be more disperse than the model's prediction. All in all, the model performs considerably similar to the humans in all but one of the 6 words shown: when asked to chose the color evoked by the word \texttt{EGG} human subjects tend to select colors related to the yolk (yellow-like), while the model deviates towards the white-blues. One may think that the model is choosing the color most similar to \emph{white} but, we found that it tends to choose blue-like colors when it is not sure about the correct color. 

\begin{table}[h!]
    \scriptsize
  \caption{P-values for Hotelling test on model vs human responses. {\color{myred}Red}: human/CLIP differ. \textbf{Bold}:  words in Fig. \ref{fig:MediasHC}.}
  \label{tab:hotelling}
  \begin{tabular}{@{}lc | lc | lc@{}} 
    \toprule
    \textbf{Word} & \textbf{p-value} & \textbf{Word} & \textbf{p-value} & \textbf{Word} & \textbf{p-value} \\
    \midrule \midrule
    {\color{myred} RAGE} & {\color{myred} 3.22e-10} & {\color{myred} MUD} & {\color{myred} 1.49e-09} & {\color{myred} APPLE} & {\color{myred} 6.69e-09} \\
    {\color{myred} DISGUST} & {\color{myred} 1.03e-05} & {\color{myred}  \textbf{EGG}} & {\color{myred}  \textbf{3.47e-05}} & {\color{myred} PEACE} & {\color{myred} 1.90e-04} \\
    {\color{myred} HAIR} & {\color{myred} 6.94e-04} & {\color{myred} BARBIE} & {\color{myred} 1.77e-03} & {\color{myred} WHALE} & {\color{myred} 2.39e-03} \\
    SEA & 4.73e-03 & SHAME & 8.09e-03 & \textbf{SAPPHIRE} & \textbf{0.0162} \\
    AMATIST & 0.0189 & SUNFLOWER & 0.0202 & SAND & 0.0224 \\
    \textbf{BASIL} & \textbf{0.0795} & TRACTOR & 0.0795 & POND & 0.0876 \\
    \textbf{LAVENDER} & \textbf{0.106} & \textbf{LEMON} & \textbf{0.140} & BLUSH & 0.147 \\
    KIWI & 0.169 & \textbf{BLOOD} & \textbf{0.186} & SALMON & 0.248 \\
    TOMATO & 0.264 & BANANA & 0.329 & PUMPKIN & 0.342 \\
    GRASS & 0.372 & DANGER & 0.508 & CUCUMBER & 0.620 \\
    LIME & 0.622 & FEMINISM & 0.732 & AUBERGINE & 0.766 \\
    SKIN & 0.843 &  &  &  &  \\ 
    \bottomrule
  \end{tabular}
\end{table}

Table \ref{tab:hotelling} shows that, for the 34 words considered in our experiment, CLIP doesn't answers like a human in 9 cases (words in red). This results in an error rate of $26\%$. 
We found that humans preferred red-like colors for the word \texttt{APPLE}, while the model prefers green-like colors. At the same time, we found that CLIP shares a western-centric bias for clearer \texttt{SKIN} colors. This suggest that our approach to testing models may indicate certain biases in the dataset that would be hard to capture with more traditional testing strategies.
A similar thing could be said about the different levels of abstraction that different words can represent. As an example, words like \texttt{AMATIST}, \texttt{BLOOD} or \texttt{BANANA} represent concise elements, while words like \texttt{RAGE}, \texttt{DISGUST} and \texttt{PEACE} represent feelings or concepts that go beyond tangible things. It is no surprise that the artificial model performs worse at these higher abstraction levels. It is also interesting to note that, 
humans show agreement in their answers. 

\section{Conclusions}
\label{sec:conclusion}

In this work we explore a new way of assessing the alignment between artificial models and humans by using a particular board game (Hues \& Cues) that puts to test the relations between colors and words that humans have developed. This relations span a wide range of abstraction levels and complexities and can even unveil cultural biases in the models. By doing so, we were able to see that CLIP has some issues recognizing more abstract concepts like \texttt{RAGE} and \texttt{DISGUST}, shares western-centric biases with respect to skin color, and has also a predilection for a different color of apples with respect to humans. We consider this the main contribution of our work: presenting board games as a useful tool to evaluate the biases and human-like behaviors of models. This is not done with the intent of ditching currently used benchmarks but with the objective of broadening our tools because, as models become more complicated and presumably more human, we will need new tools to evaluate their emerging capabilities.


\section{Acknowledgements}

This work was supported in part by MICIIN/FEDER/UE under Grant PID2020-118071GB-I00  and PDC2021-121522-C21, in part by Generalitat Valenciana under Projects GV/2021/074, CIPROM/2021/056 and  CIAPOT/2021/9, and Grant CIACIF/2023/223. Some computer resources were provided by Artemisa, funded by the European Union ERDF and Comunitat Valenciana as well as the technical support provided by the Instituto de Física Corpuscular, IFIC (CSIC-UV). Special thanks to María Alabau Bosque for collaborating in designing tasks in the figures shown and the presented poster.

\bibliographystyle{ccn_style}

\setlength{\bibleftmargin}{.125in}
\setlength{\bibindent}{-\bibleftmargin}

\bibliography{ccn_style}

\begin{thebibliography}{}

\bibitem [\protect \citeauthoryear {%
Brown%
\ \BBA {} Lindsey%
}{%
Brown%
\ \BBA {} Lindsey%
}{%
{\protect \APACyear {2023}}%
}]{%
brown_color_2023}
\APACinsertmetastar {%
brown_color_2023}%
\begin{APACrefauthors}%
Brown, A\BPBI M.%
\BCBT {}\ \BBA {} Lindsey, D\BPBI T.%
\end{APACrefauthors}%
\unskip\
\newblock
\APACrefYearMonthDay{2023}{{\APACmonth{09}}}{}.
\newblock
{\BBOQ}\APACrefatitle {The color communication game} {The color communication game}.{\BBCQ}
\newblock
\APACjournalVolNumPages{Scientific Reports}{13}{1}{16006}.
\newblock
\begin{APACrefURL} \url{https://doi.org/10.1038/s41598-023-42834-3} \end{APACrefURL}
\newblock
\begin{APACrefDOI} \doi{10.1038/s41598-023-42834-3} \end{APACrefDOI}
\PrintBackRefs{\CurrentBib}

\bibitem [\protect \citeauthoryear {%
Hotelling%
}{%
Hotelling%
}{%
{\protect \APACyear {1931}}%
}]{%
Hotelling}
\APACinsertmetastar {%
Hotelling}%
\begin{APACrefauthors}%
Hotelling, H.%
\end{APACrefauthors}%
\unskip\
\newblock
\APACrefYearMonthDay{1931}{}{}.
\newblock
{\BBOQ}\APACrefatitle {The Generalization of Student's Ratio} {The generalization of student's ratio}.{\BBCQ}
\newblock
\APACjournalVolNumPages{The Annals of Mathematical Statistics,}{2}{}{360-378}.
\PrintBackRefs{\CurrentBib}

\bibitem [\protect \citeauthoryear {%
Kay%
\ \BBA {} Cook%
}{%
Kay%
\ \BBA {} Cook%
}{%
{\protect \APACyear {2023}}%
}]{%
kay2023world}
\APACinsertmetastar {%
kay2023world}%
\begin{APACrefauthors}%
Kay, P.%
\BCBT {}\ \BBA {} Cook, R\BPBI S.%
\end{APACrefauthors}%
\unskip\
\newblock
\APACrefYearMonthDay{2023}{}{}.
\newblock
{\BBOQ}\APACrefatitle {World color survey} {World color survey}.{\BBCQ}
\newblock
\BIn{} \APACrefbtitle {Encyclopedia of Color Science and Technology} {Encyclopedia of color science and technology}\ (\BPGS\ 1601--1607).
\newblock
\APACaddressPublisher{}{Springer}.
\PrintBackRefs{\CurrentBib}

\bibitem [\protect \citeauthoryear {%
Lindsey%
\ \BBA {} Brown%
}{%
Lindsey%
\ \BBA {} Brown%
}{%
{\protect \APACyear {2021}}%
}]{%
lindsey_lexical_2021}
\APACinsertmetastar {%
lindsey_lexical_2021}%
\begin{APACrefauthors}%
Lindsey, D\BPBI T.%
\BCBT {}\ \BBA {} Brown, A\BPBI M.%
\end{APACrefauthors}%
\unskip\
\newblock
\APACrefYearMonthDay{2021}{}{}.
\newblock
{\BBOQ}\APACrefatitle {Lexical Color Categories} {Lexical color categories}.{\BBCQ}
\newblock
\APACjournalVolNumPages{Annual Review of Vision Science}{7}{Volume 7, 2021}{605-631}.
\PrintBackRefs{\CurrentBib}

\bibitem [\protect \citeauthoryear {%
Malo%
\ \BBA {} Luque%
}{%
Malo%
\ \BBA {} Luque%
}{%
{\protect \APACyear {2002}}%
}]{%
Colorlab02}
\APACinsertmetastar {%
Colorlab02}%
\begin{APACrefauthors}%
Malo, J.%
\BCBT {}\ \BBA {} Luque, M.%
\end{APACrefauthors}%
\unskip\
\newblock
\APACrefYearMonthDay{2002}{}{}.
\newblock
{\BBOQ}\APACrefatitle {{ColorLab: A Matlab Toolbox for Color Science and Calibrated Color Image Processing}} {{ColorLab: A Matlab Toolbox for Color Science and Calibrated Color Image Processing}}.{\BBCQ}
\newblock
\APACjournalVolNumPages{Univ. Valencia. https://isp.uv.es/code/vision\_and\_color/colorlab}{}{}{}.
\PrintBackRefs{\CurrentBib}

\bibitem [\protect \citeauthoryear {%
Radford%
\ \protect \BOthers {.}}{%
Radford%
\ \protect \BOthers {.}}{%
{\protect \APACyear {2021}}%
}]{%
clip}
\APACinsertmetastar {%
clip}%
\begin{APACrefauthors}%
Radford, A.%
\BCBT {}\ \BOthersPeriod {.}
\end{APACrefauthors}%
\unskip\
\newblock
\APACrefYearMonthDay{2021}{}{}.
\newblock
{\BBOQ}\APACrefatitle {Learning transferable visual models from natural language supervision} {Learning transferable visual models from natural language supervision}.{\BBCQ}
\newblock
\BIn{} \APACrefbtitle {International conference on machine learning} {International conference on machine learning}\ (\BPGS\ 8748--8763).
\PrintBackRefs{\CurrentBib}

\end{thebibliography}

\begin{figure*}[]
	\centering
	\includegraphics[width=1\linewidth]{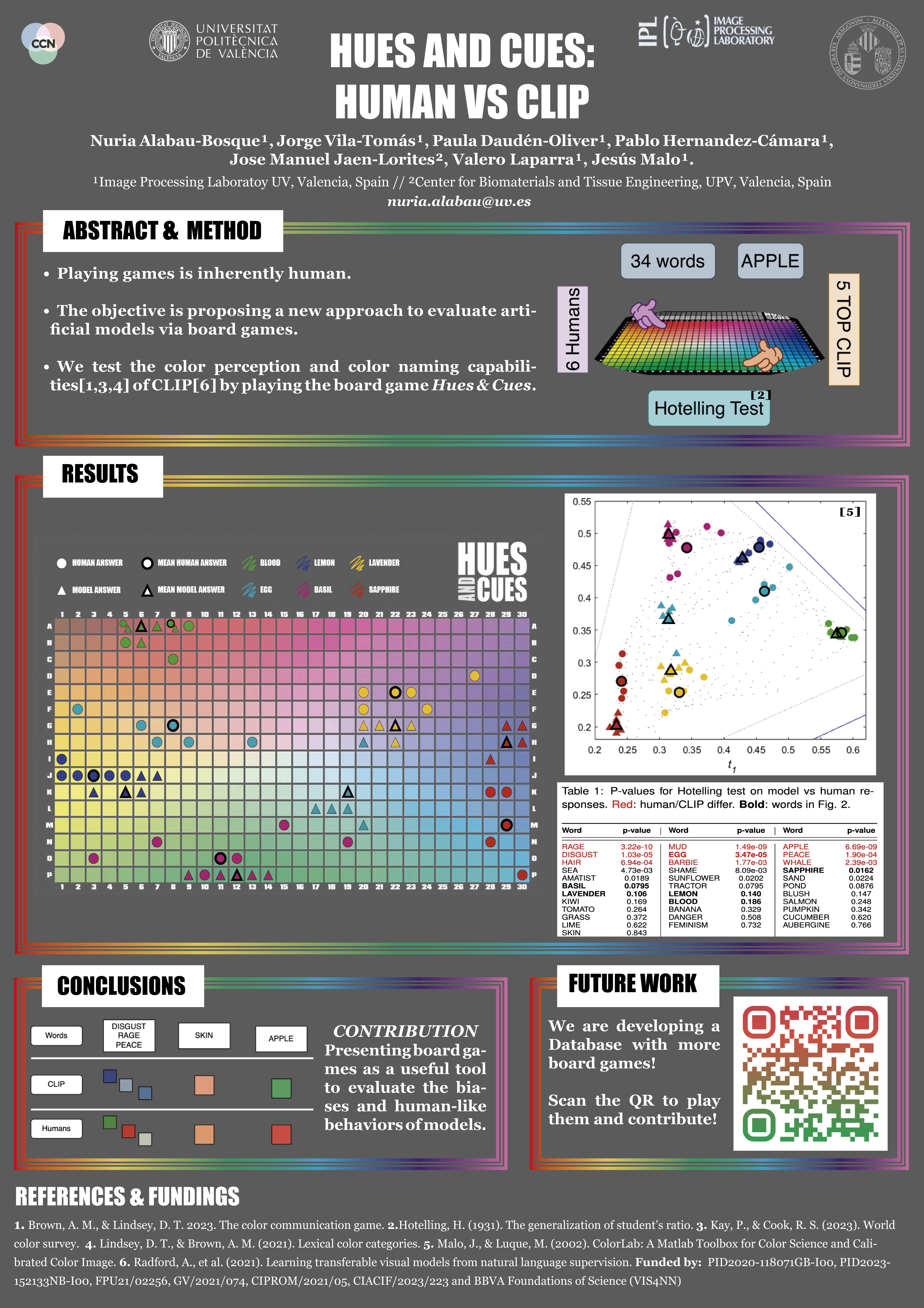}
	\label{fig:Poster}
    \end{figure*}

\end{document}